\documentclass[sigconf]{acmart}

\AtBeginDocument{%
  \providecommand\BibTeX{{%
    \normalfont B\kern-0.5em{\scshape i\kern-0.25em b}\kern-0.8em\TeX}}}

\acmConference[ASP-DAC 2021]{ACM Asia South Pacific Design Automation Conference}{June 18--21, 2021}{Tokyo, Japan}

\usepackage{subfig}
\usepackage{tablefootnote}
\usepackage{footnote}
\usepackage{threeparttable}

\begin{document}

\title{On Designing Computing Systems for Autonomous Vehicles: a PerceptIn Case Study}

\author{Bo Yu}
\email{bo.yu@perceptin.io}
\affiliation{
  \institution{PerceptIn Inc.}
}

\author{Jie Tang}
\authornote{corresponding author}
\email{cstangjie@scut.edu.cn}
\affiliation{%
  \institution{South China University of Technology}
}

\author{Shaoshan Liu}
\email{shaoshan.liu@perceptin.io}
\affiliation{%
  \institution{PerceptIn Inc.}
}



\begin{abstract}
  PerceptIn develops and commercializes autonomous vehicles for micromobility around the globe. This paper makes a holistic summary of PerceptIn's development and operating experiences. This paper provides the business tale behind our product, and presents the development of the computing system for our vehicles. We illustrate the design decision made for the computing system, and show the advantage of offloading localization workloads onto an FPGA platform.
\end{abstract}

\ccsdesc[500]{Computer systems organization~Robotics}    
 
\keywords{autonomous vehicle, FPGA, localization}


\maketitle

\begin{figure*}[t]
\centering 
\begin{minipage}[b]{0.24\textwidth} 
	\centering 
	\includegraphics[width=\columnwidth]{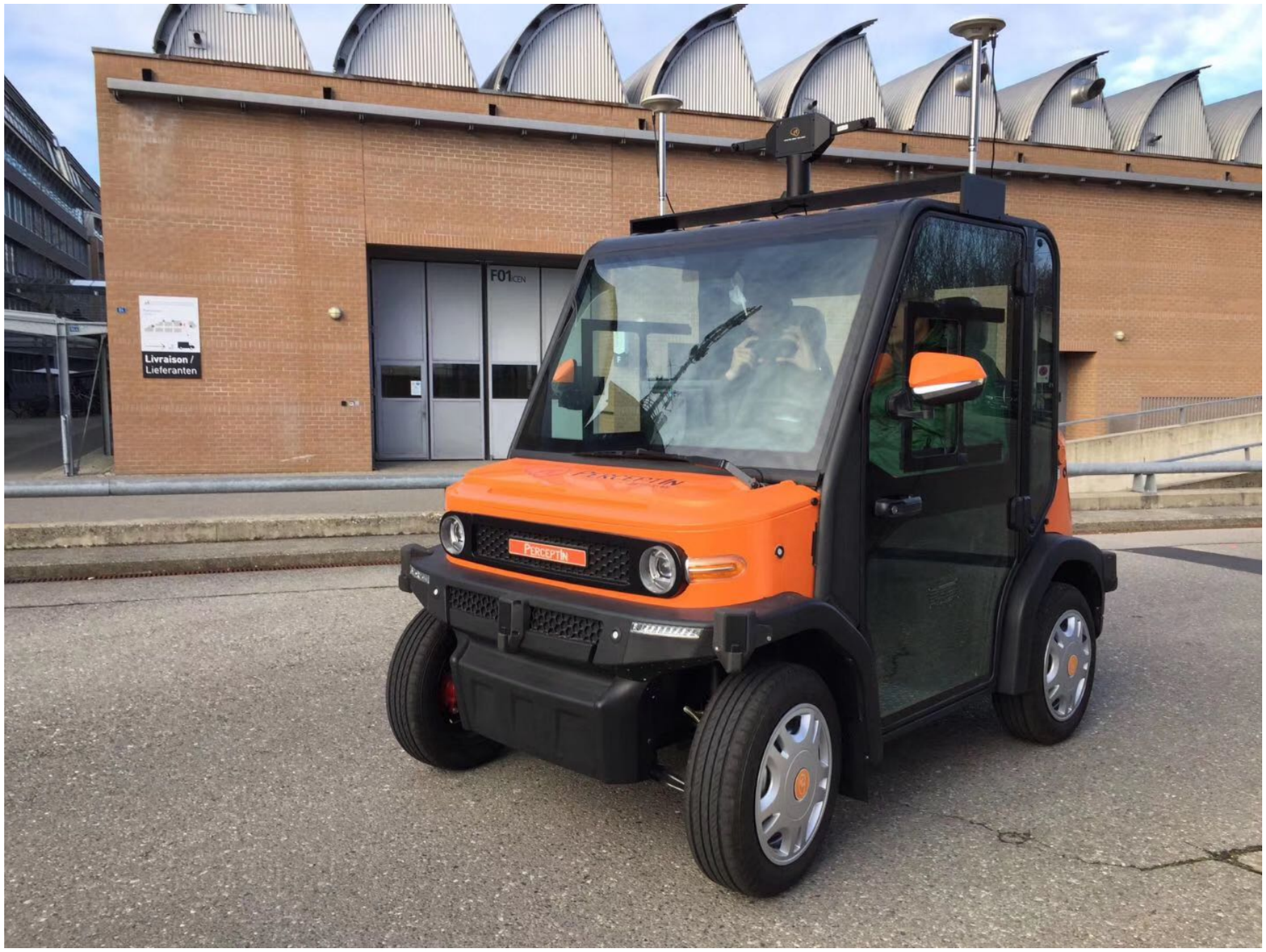}
	\caption{DragonFly pod in the field.}
	\label{fig:car}
\end{minipage}
\hspace{2pt}
\begin{minipage}[b]{0.74\textwidth} 
	\centering 
	\includegraphics[width=\columnwidth]{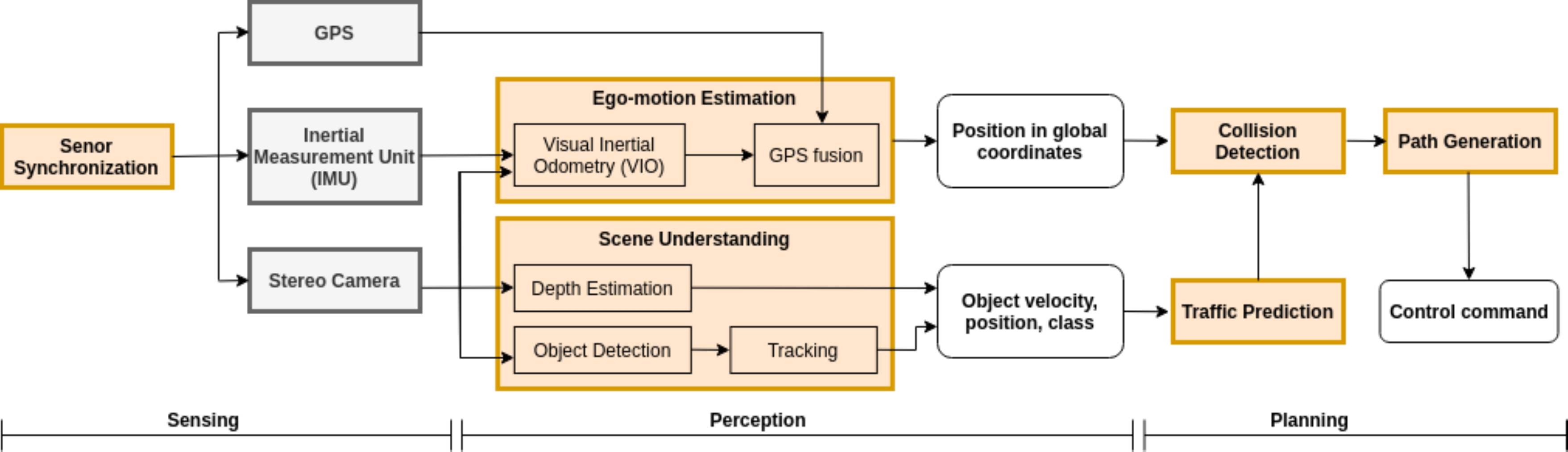}
	\caption{Processing pipeline of our autonomous vehicles.}
	\label{fig:pipeline}
\end{minipage}
\end{figure*}


\section{Introduction}
PerceptIn was established in 2016 to develop visual perception technologies for autonomous vehicles and robots. Since its inception, PerceptIn has successfully attracted over \$10 million of venture capital funding, from Walden International, Matrix Partners, and Samsung Ventures \cite{liu2020critical}.  PerceptIn is an international technology startup with operations in the U.S., Japan, Europe, and Asia. PerceptIn consists of over 30 researchers and engineers and 10 business professionals. The business professionals are responsible for business development in different markets and gather feedback for the company's R\&D efforts, whereas the engineers and researchers are responsible for developing cutting edge autonomous driving technologies. In the past three years, PerceptIn has generated over 20 U.S. patents and over 100 international patents, as well as numerous research papers. 

In 2017, PerceptIn decided to develop low speed autonomous vehicles to serve the micromobility market, as micromobility is a rising transport mode wherein lightweight vehicles cover short trips that massive transit ignore \cite{liu2020autonomous}. According to US Department of Transportation, 60\% of vehicle traffic is attributed to trips under 5 miles \cite{santos2011summary}. Transportation needs in short trips are disproportionally under-served by current mass transit systems due to high cost, which affects the society profoundly. Micromobility bridges transit services and communities' needs, driving the rise of Mobility-as-a-Service.

Based on internal business analysis, if PerceptIn can provide low-speed autonomous vehicles under \$70,000 per unit, PerceptIn could generate a reasonable return-on-investment for PerceptIn's customers, the autonomous vehicle operators. However, the \$70,000 price tag also imposes very strict and challenging constraints on the design of low-speed autonomous vehicles. In detail, we have to break down the \$70,000 into Non-recurring engineering (NRE) cost such as research and development, recurring costs such as the cost of the chassis, the cost of drive-by-wire conversion (meaning to convert a traditional vehicle into one that can be controlled by computers), the cost of sensors, the cost of integration, the cost of customer service, and finally the cost of the computing system \cite{liu2020engineering}.

As the on-vehicle computing system is a significant contributor to cost and power consumption, PerceptIn conducted a study on autonomous driving computing systems \cite{liu2017computer}\cite{liu2019edge}\cite{liu2020pirt}, PerceptIn concluded that computing is the bottleneck for the commercial deployment of autonomous vehicles, and PerceptIn needed a computing system that is reliable, affordable, high-performance, and energy efficient. Most importantly, PerceptIn needed a solution that is cost effective and has a short time-to-market. PerceptIn faced several options:

Option one, Optimization of commercial off-the-shelf mobile System-on-Chip (SoC) computing systems: This approach brings several benefits, first, since mobile SoCs have reached economies of scale, it would have been most beneficial for PerceptIn to build its technology stack on affordable, backward-compatible computing systems. Second, PerceptIn's vehicles target micromobility with limited speed, similar to mobile robots, for which mobile SoCs have been demonstrated before.  However, an extensive study is required to fully understand mobile SoCs' suitability for autonomous driving, this may delay PerceptIn's product launch by six months.   

Option two, Procurement of specialized autonomous driving computing systems: there were commercial computing platforms specialized for autonomous driving, such as those from NXP, MobilEye, and Nvidia. They are mostly Application-Specific Integrated Circuit (ASIC) based chips that provide high performance at a much higher cost. For instance, the first-generation of Nvidia PX2 system costs over \$10,000. Besides the cost issue, these computing systems mostly accelerate only the perception function in autonomous driving, whereas PerceptIn require a system that optimizes the end-to-end performance. So we soon concluded that this option was not viable.

Option three, Development of proprietary autonomous driving computing systems: developing a proprietary computing system guarantees that PerceptIn have the most suitable system for PerceptI's customers and for its workloads, but also means that PerceptIn need to invest a significant amount of financial and personnel resources on this project. Also, the investment does not guarantee the success of this project. It is a huge and risky bet for a startup like PerceptIn. 

After an unsuccessful exploration with option one, starting in early 2018, PerceptIn decided to move forward with option three and PerceptIn thus formed a team to develop the FPGA-based DragonFly computing system \cite{fang2018dragonfly+}\cite{fang2017fpga}\cite{liu2020pi}\cite{tang2018pi}\cite{yu2020building}. Option three was a huge success, today all autonomous vehicles shipped by PerceptIn are empowered by PerceptIn's proprietary DragonFly computing system. In this paper, from a technical perspective we explained how we designed the PerceptIn's DragonFly computing systems, and the design trade-offs that we faced. 

For the rest of this paper, we first introduce the sensor suites of our vehicle (Sec. \ref{sec:veh}), and then present the end-to-end processing pipeline exercised by the on-vehicle computing systems (Sec. \ref{sec:alg}). Based on the processing pipeline, we then describe the on-vehicle computing system design (Sec. \ref{sec:sov}). To reduce the end-to-end computing latency, we offload the localization to an FPGA platform.
Sec. \ref{sec:fpga} presents FPGA based vSLAM (visual simultaneous localization and mapping) front-end and back-end systems designed for our vehicle.

\section{Our Vehicles and Sensor Suite}
\label{sec:veh}

We first briefly introduce our product and sensor suite deployed on our vehicle in order to provide contexts for the rest of the paper.

Our vehicles are electric cars that are powered by batteries. We provide two autonomous vehicles designs: 2-seater pods targeting private transportation experiences, and 8-seater shuttles targeting public autonomous driving transportation services. Both designs are capped at 20 mph to suit the unique needs of micromobility. 

The unique scenarios of micromobility let us build affordable autonomous vehicles. 
We use cameras as the major sensor for the perception and localization, and develop the sensor suite, Dragonly, for our vehicle. DragonFly integrates two sets of stereo cameras, which cover forward and backward regions respectively, and an IMU (Inertial Measurement Unit) for pose estimation. Besides cameras, radars and sonars are used to cover the short range regions and to provide the 360 degree of perception coverage.

\section{Algorithm Pipeline}
\label{sec:alg}

\textbf{Processing pipeline.} Fig. \ref{fig:pipeline} shows the block diagram of the processing pipeline in our vehicle, which consists of three parts: sensing, perception and planning. The sensing module bridges sensors and the computing system. It synchronizes various sensor samples for the downstream perception module, which performs two fundamental tasks: 1) locating the vehicle itself in a global map and 2) understanding the surroundings through depth estimation and object detection. The planning module uses the perception results to devise a drivable route, and then converts the planed path into a sequence of control commands, which will drive the vehicle along the path. The control commands are sent to the vehicle's Engine Control Unit (ECU) via the CAN bus interface.


\textbf{Algorithm.}
Our localization module is based on Visual Inertial Odometry algorithms \cite{qin2018vins,sun2018msckf}, which fuses camera images, IMU and GPS samples to estimate the vehicle pose in the global map. The depth estimation employs traditional stereo vision algorithms, which calculates depths according to the principal of triangulation \cite{szeliski2010computer}. In particular, our method is based on the classic ELAS algorithm, which uses hand-crafted features \cite{Elas2010Geiger}. While DNN models for depth estimation exist, they are orders of magnitude more compute-intensive than non-DNN algorithms\cite{feng2019stereo} while providing only marginal accuracy improvements to our use-cases. 

We detect objects using DNN models. Object detection is the only task in our current pipeline where the accuracy provided by deep learning justifies the overhead. An object, once detected, is tracked across time until the next set of detected objects are available. The DNN models are trained regularly using our field data. As the deployment environment can vary significantly, different models are specialized/trained using the deployment environment-specific training data. We use the Kernelized Correlation Filter (KCF) \cite{henriques2014high} to track detected objects. The planning algorithm we used is basd on Model Predictive Control (MPC)\cite{kelly2013mobile}.

\textbf{Task-Level Parallelism.}
As shown in Fig. \ref{fig:pipeline}, sensing, perception and planning are serialized. They are all on the critical path of the end-to-end latency. We pipeline the three modules to improve the throughput. Different sensor processes (e.g., IMU vs. camera) are independent. Within perception, localization and scene understanding are independent and could execute in parallel. 

While there are multiple tasks within scene understanding, they are mostly independent with the only exception that object tracking must be serialized with object detection. The task-level parallelisms influence how the tasks are mapped to the hardware platform as we will discuss later.

\begin{figure}[t]
\centering
\subfloat[Latency comparison.]
{
  \includegraphics[trim=0 0 0 0, clip, width=\columnwidth]{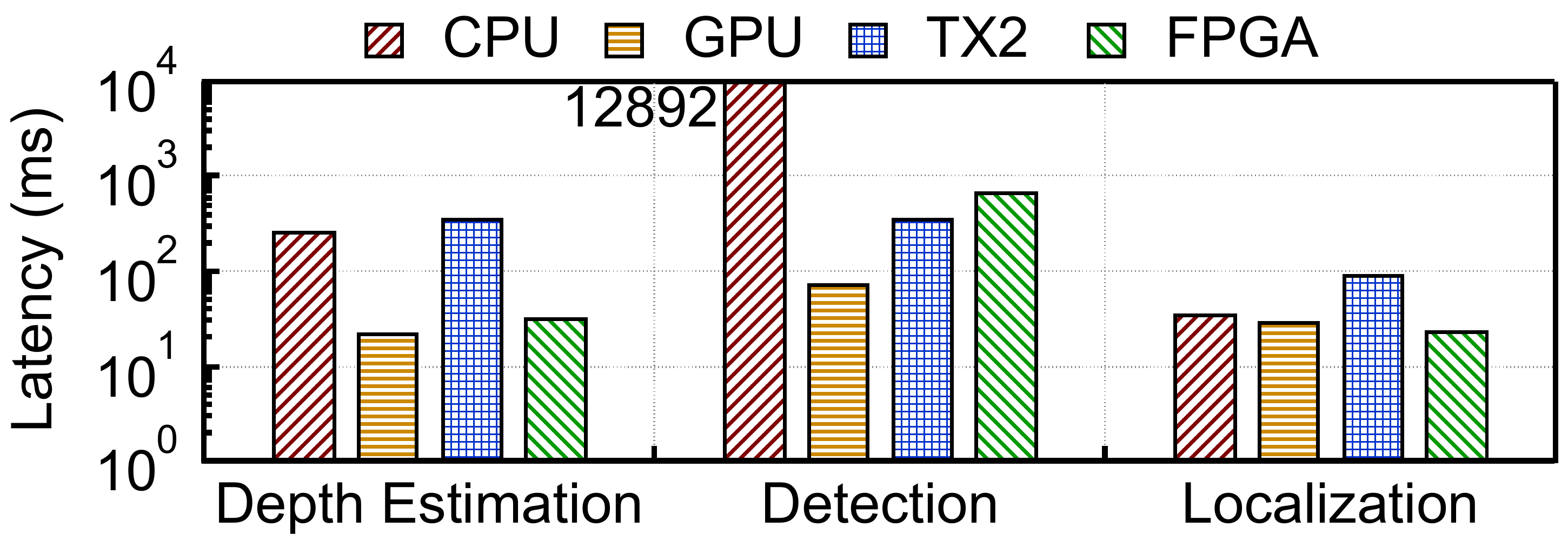}
  \label{fig:dse_l}
}
\\
\subfloat[Energy consumption comparison.]
{
  \includegraphics[trim=0 0 0 0, clip, width=\columnwidth]{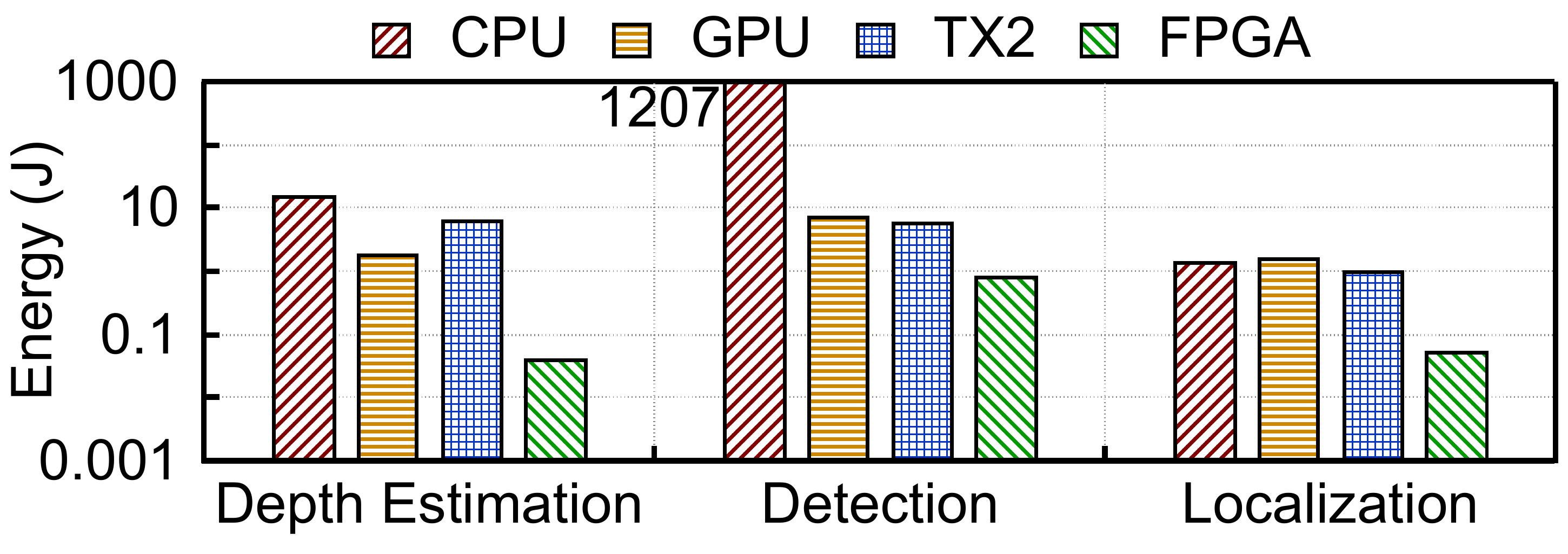}
  \label{fig:dse_e}
}
\caption{Performance and energy comparison of four different platforms running three perception tasks.}
\label{fig:dse}
\vspace{-10pt}
\end{figure}

\section{On-Vehicle Computing System}
\label{sec:sov}

\subsection{Hardware Design Space Exploration}
This section shows the hardware design space explored during our development. 

\textbf{Moible SoCs.}
Since our vehicles targets micromobility with low speed, which has much in common with mobile robots, high-end mobile SoCs that have been demonstrated on robots are initially explored. However, we found that mobile SoCs are ill-suited for autonomous driving as the main computing device. The major reason is that the compute capability of mobile SoCs is too low for realistic end-to-end autonomous driving workloads.

Fig. \ref{fig:dse} shows the latency and energy consumption of three perception tasks---depth estimation, object detection, and localization---on an Intel Coffee Lake CPU (3.0 GHz, 9 MB LLC), Nvidia GTX 1060 GPU, and Nvidia TX2, which represents today's high-end mobile SoCs. 
On TX2, we use the Pascal GPU for depth estimation and object detection, and use the ARM Cortex-A57 CPU (with SIMD capabilities) for localization, which is ill-suited for GPU due to the lack of massive parallelisms. Fig. \ref{fig:dse_l} shows that TX2 is much slower than the GPU, leading to a cumulative latency of 844.2 $ms$ for perception alone. Fig. \ref{fig:dse_e} shows that TX2 has only marginal, sometimes even worse, energy reduction compared to the GPU due to the long latency.

On the other hand, although rich general IO ports are available on mobile SoCs, sensor processing supports in hardware, which is critical for multi-sensor fusion algorithm, are missed. For example, autonomous vehicles require very precise and clean sensor synchronization for accurate pose estimations, which mobile SoCs do not directly provide.

\textbf{FPGA vs GPU.}
We implement depth estimation, YOLO \cite{redmon2018yolov3} and ORB front-end on a Xilinx Zynq UltraScale+ FPGA and Nvidia GTX 1060. For FPGA design, we write RTL codes and use Xinlinx Vivado for implementation. For GPU design, we implement depth and ORB front-end algorithm using CUDA.  

Fig. \ref{fig:dse_e} compares the latency of the perception tasks on the FPGA with the GPU. The GPU is faster than the FPGA on depth estimation and object detection mainly due to the highly parallel and regular operations in these image processing pipeline. The FPGA is faster than the GPU on localization, which inherently lacks massive parallelisms and is more lightweight than other tasks.

\begin{figure}[t]
\centering
\includegraphics[width=1\columnwidth]{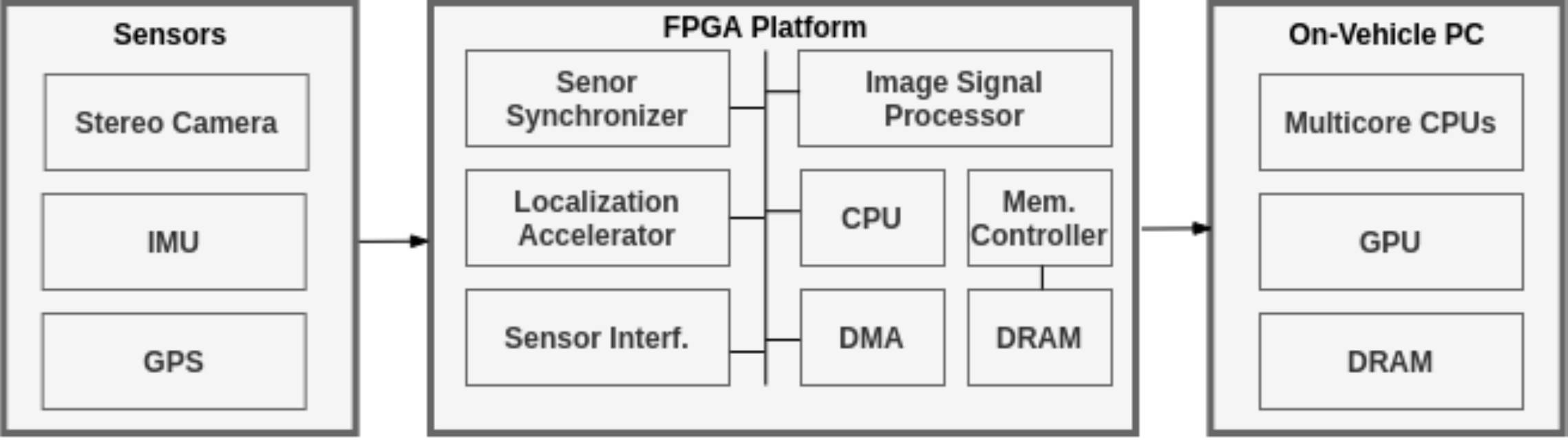}
\caption{The computing system in our autonomous vehicle.}
\label{fig:hwarch}
\end{figure} 

\subsection{Hardware Architecture}
The results from our hardware exploration motivate the on-vehicle computing system design. Fig. \ref{fig:hwarch} illustrates an overview of the on-vehicle computing system. It includes sensors and a server+FPGA heterogeneous platform. Components that are not on the critical path, such as Radar and Sonar, are not shown here. 

Considering the cost, compute requirements and power budget, our computing platform is composed of a Xilinx Zynq Ultrascale+ FPGA and an on-vehicle PC equipped with an Intel Coffe Lake CPU and an Nvidia GTX 1060 GPU. The PC is the main computing platform, while the FPGA plays a critical role, which bridges sensors and the PC, and provides an acceleration platform. 

\textbf{Algorithm mapping.}
We map sensing to the Zynq FPGA platform, which essentially acts a sensor hub. It processes sensor data and transfers sensor data to the PC for subsequent processing. The Zynq FPGA hosts an ARM-based SoC and runs a full-fledged Linux OS, on which we develop sensor processing and data transfer pipelines.

We assign the planning tasks to the CPU of the on-vehicle server, for two reasons. First, the planning commands are sent to the vehicle through the CAN bus; the CAN bus interface is simply more mature on high-end servers than embedded FPGAs. Executing the planning module on the server greatly eases deployment. Second, we optimize our planning algorithm for micromobility, which leads to an extreme lightweight algorithm. We use a lane-level map instead of the grid high precision map. The search and optimization space of feasible trajectories is greatly reduced. On our computing system, the planning only contributes to about 1\% of end-to-end latency.  

Perception tasks include scene understanding (depth estimation and object detection/tracking) and localization, which are independent and, therefore, the slower one dictates the overall perception latency. Our design offloads localization onto the FPGA while leaving other perception tasks on the GPU. This partitioning also frees more GPU resources for depth estimation and object detection, further reducing latency.

When both scene understanding and localization execute on the GPU, they compete for resources and slow down each other. Scene understanding takes 120 $ms$ and dictates the perception latency. When localization is offloaded to the FPGA, 
the localization and scene understanding are executed in parallel. Since scene understanding is more computationally intensive, scene understanding is still on the critical path, but its latency reduces to 77 $ms$. Overall, the perception latency improves by 1.6 $\times$.

\section{FPGA Based Localization System}
\label{sec:fpga}

This section presents an FPGA based localization front-end and back-end designed for our vehicles, which accelerates stereo ORB SLAM \cite{mur2017orb}. 

Modern FPGA platforms are built with rich and mature sensor interfaces and image processing IPs. As a result, our FPGA platform directly interfaces with the stereo cameras on DragonFly, of which image resolution is 1280$\times$720 pixels. We designed FPGA circuits to synchronize two cameras by triggering cameras at the same time. Sensor synchronization is crucial to perception algorithms, which fuse multiple sensors. Out-of-sync sensor data is detrimental to perception. For stereo vision, even small temporal offsets will lead to large depth estimation error, which further compromise the accuracy of the localization algorithm. 

\subsection{Algorithm}
Our SLAM system includes two components: the front-end and the back-end. The front-end extracts image features and associates features in consecutive frames to physical landmarks. It incrementally deduces the robot motion by applying geometry constraints on the associated sensory observations. The back-end tries to minimize errors introduced by the process and measurement noises by performing optimizations on a batch of observed landmarks and tracked poses.

\begin{figure}[t]
\centering
\includegraphics[width=0.9\columnwidth]{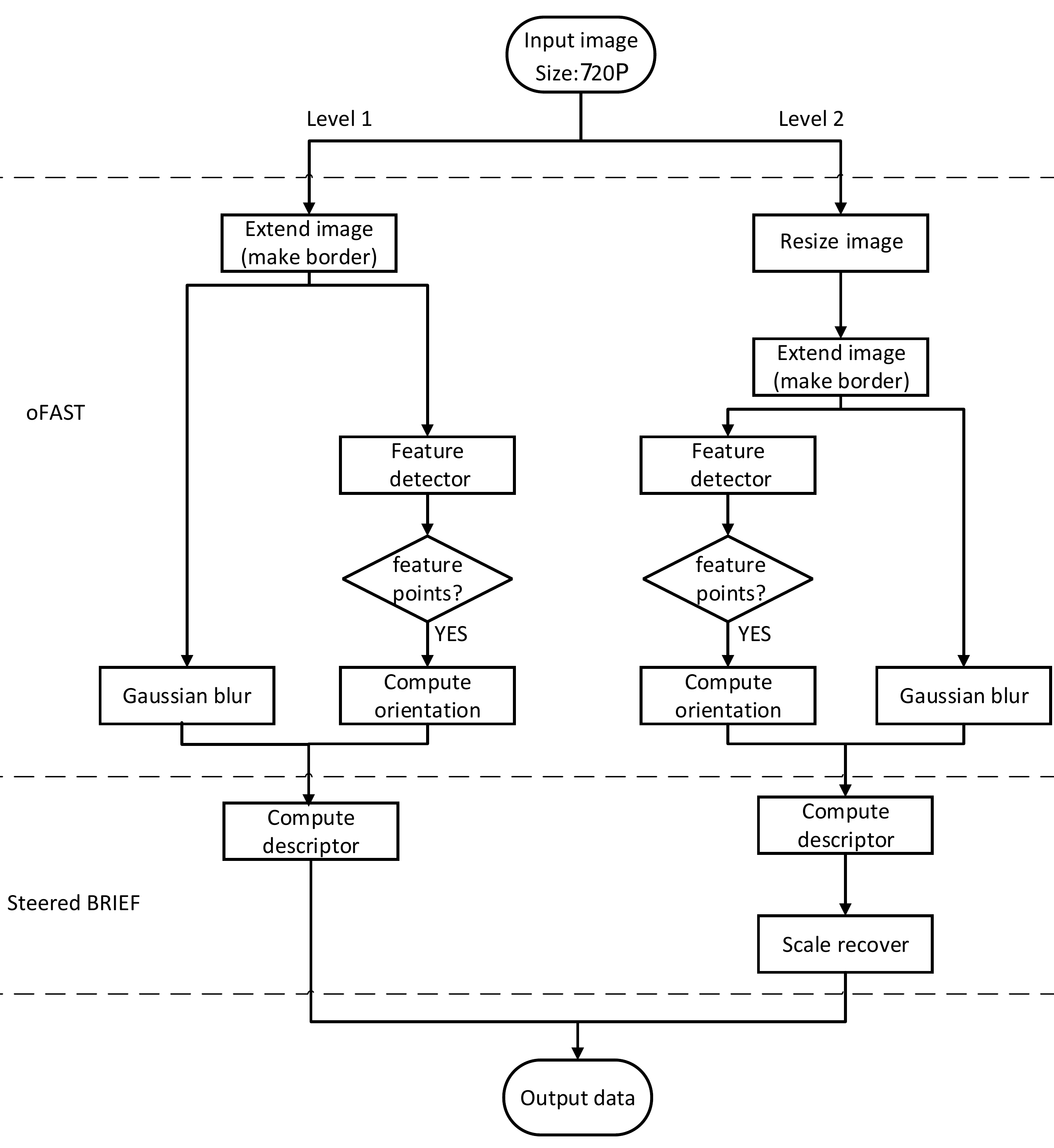}
\caption{ORB feature extraction pipeline.}
\label{fig:orb_d}
\vspace{-5mm}
\end{figure} 

\subsubsection{Front-end} 
The front-end consists of two parts: ORB feature extraction and feature association.
\begin{itemize}
  \item \textbf{Feature Extraction.} The front-end first extract key points on 2D images, which correspond to salient landmarks in the 3D world. The pipeline of ORB feature extraction is shown in Fig. \ref{fig:orb_d}. The input is a gray image. An n-level image pyramid is constructed to make the feature scale invariant. In our design, n is 2, which makes a trade-off between algorithm accuracy and computational intensity. Key points are detected using FAST feature on pyramid images. If a feature is detected, the orientation of the feature point is calculated to make the feature rotation invariant. Feature descriptors are computed on Gaussian blurred images using steered BRIEF descriptor. Lastly, feature descriptors of resized images are scaled back to the original image. 
  \item \textbf{Feature association.} This module compares features' descriptor in two images and associates features that correspond to the same landmarks. We use 256 bits descriptors for features, and use Hamming distance between descriptors as the matching criteria. The pair of feature, of which Hamming distance is below a threshold, is considered as a matched pair. To improve matching accuracy, SAD (sum of absolute differences) rectification is applied. Comparing all the pairs of features in two images to find matched features is computationally expensive. The relative rotation and translation between two images provide geometry constraints on the corresponding features in the image pair, which limit the searching space on a band region on the image. We use IMU data to get the movement between two consecutive images, and get relative rotation and translation of stereo cameras from the sensors' calibration process \cite{kalibr2013}. In the system, the relative rotation and translation between images pairs and its constraints are employed to find matched features.	  
\end{itemize}

\subsubsection{Back-end} 
The front-end produces initial estimations of landmark positions and camera poses. The back-end further refines camera trajectories and 3D structures. We use bundle adjustment as the back-end method. 

Bundle adjustment \cite{Triggs1999Bundle} is actually an optimization problem, which is formulated by Eq.~\eqref{eq:costfunction}. $P(\boldsymbol{p_i}, \boldsymbol{c_j})$ is the camera projection model that projects a 3D point in the world coordinate, $\boldsymbol{p_i}$, to a 2D point on the image plane, $j$. $\boldsymbol{c_j}$ represents the parameters of the $j-th$ camera pose. $\boldsymbol{o_{ij}}$ is the feature corresponding to the $i$-th point on the $j$-th image. $\epsilon_{ij}$ is the discrepancy between the observation and the projection function. 

\begin{figure}[htbp]
\centerline{\includegraphics[width=0.9\linewidth]{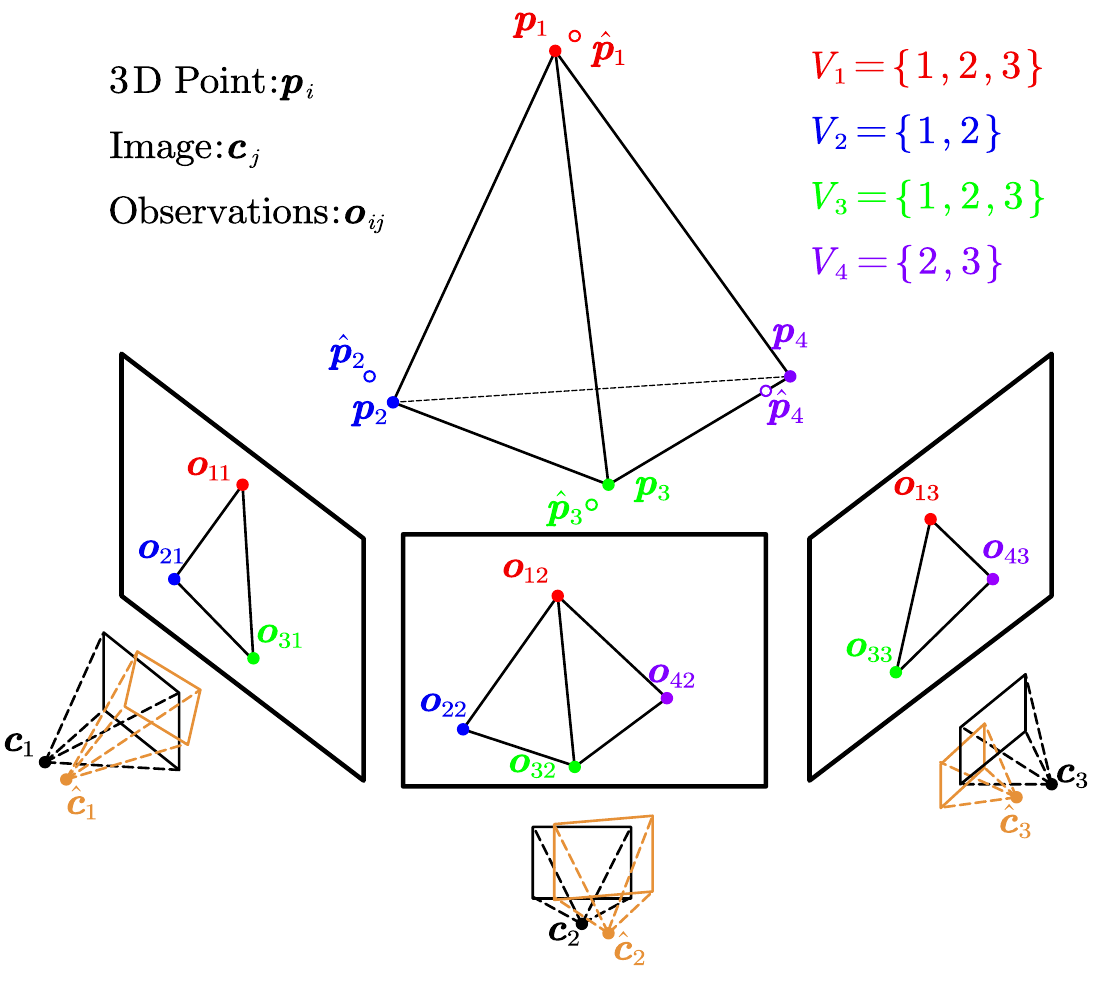}}
\caption{An example of bundle adjustment. The initial estimations of the 3D point's position and the camera's extrinsic parameter are $\hat{\boldsymbol{p_i}}$ and $\hat{\boldsymbol{c_j}}$. After bundle adjustment, optimized $\boldsymbol{p_i}$ and $\boldsymbol{c_j}$ are obtained.}
\label{fig:camera_bundle}
\vspace{-3mm}
\end{figure}

In the equation, $\sigma_{ij}$ evaluates to $1$ if the $i$-th 3D point is observed by the $j$-th camera, otherwise $0$. This formulation shows that solving the bundle adjustment problem is to determine camera parameters and 3D points' positions such that observations are closely approximated by the corresponding re-projection points. Fig.~\ref{fig:camera_bundle} is an example of bundle adjustment. The initial estimations of the 3D point's position and the camera's extrinsic parameter are $\hat{\boldsymbol{p_i}}$ and $\hat{\boldsymbol{c_j}}$. After bundle adjustment, $\boldsymbol{p_i}$ and $\boldsymbol{c_j}$ are obtained.

\begin{equation}
\small
\underset{\boldsymbol{p_{i}},\boldsymbol{c_{j}}}{\mathop{\min}}\sum_{i=1}^{a}\sum_{j=1}^{b}\sigma_{ij}\|\epsilon_{ij}\| = \underset{\boldsymbol{p_{i}},\boldsymbol{c_{j}}}{\mathop{\min}}\sum_{i=1}^{a}\sum_{j=1}^{b}{\sigma_{ij}{\parallel \boldsymbol{o_{ij}}-P\left(\boldsymbol{p_i},\boldsymbol{c_j}\right)\parallel}}
\label{eq:costfunction}
\end{equation} 

We use LM (Levenberg-Marquardt's) algorithm \cite{Lourakis2005Is}, which is widely adopted by both industry and academia, to solve bundle adjustment. The LM algorithm include five parts, which are Jacobian updates (JU), Schur elimination (SE), Cholesky factorization (CFS), cost function computation (CC) and gain ratio evaluation (GRE).
To solve non-linear optimization problems, LM algorithm iteratively use Jacobbian to linearize the problem and solve the linear equation at each iteration to obtain an update. Schur elimination is used to reduce the dimension of the linear equation, thus reduce the computational complexity. Cholesky factorization is employed to solve the linear equation to get an update of the solution. Cost function and ratio evaluation are used to determine if the update reduce the cost effectively. The update will add to the current solution if the update is effective. 

\begin{figure*}[t]
  \centering
  \includegraphics[width=\textwidth]{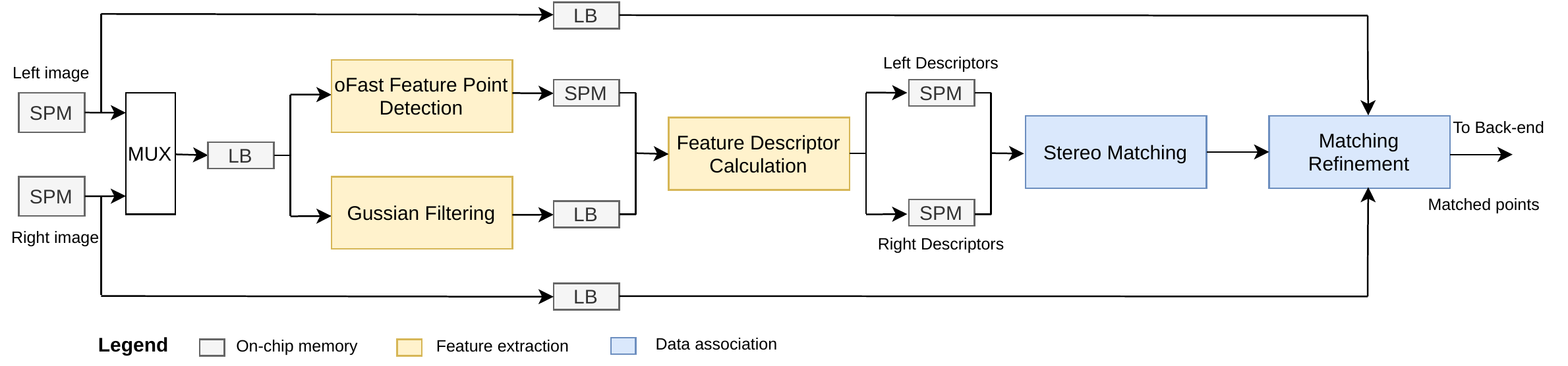}
  \caption{Front-end architecture, which consists of two parts: feature extraction and data association. On-chip memories are customized in different ways to suit different types of data reuse: line buffer (LB) supports sequential accesses, and scratchpad memory (SPM) supports irregular accesses.}
  \label{fig:fe}
\end{figure*}

\begin{table}[]
\begin{threeparttable}[b]
\caption{Comparison on resource consumption of the front-end and back-end hardware.}
\label{tab:resource}
\begin{tabular}{ccccc}
\hline
                  & LUT    & Flip-Flop & BRAM & DSP \\ \hline
Front-end         & 177.2K & 82.7K     & 785  & 109 \\
Front-end used \% & 65     & 15        & 86   & 4.3 \\ \hline
Back-end\tnote{a}         & 106.7K & 111K      & 509  & 456 \\
Back-end used \%  & 39     & 20        & 56   & 18  \\ \hline
\end{tabular}
\begin{tablenotes}
	\item[a] Back-end FPGA consists of JU, SE and CC.
\end{tablenotes}
\end{threeparttable}
\vspace{-5mm}
\end{table}

\subsection{Hardware Architecture}
We implement the front-end and back-end algorithm on a Xilinx Zyqn SoC, where computationally intensive components are implemented on FPGA logic and others are implemented on the embedded processor. 

\textbf{Front-end Architecture.} Fig. \ref{fig:fe} illustrates the front-end hardware architecture. The input (left and right) images are buffered on-chip.
The two images go through feature extraction and data association. The left and right camera streams are time-multiplexed in the feature extraction block to save hardware resources. The feature extraction block implements the ORB feature extraction in Fig. \ref{fig:orb_d}, which consists of three tasks: oFAST feature point detection (fast feature detection and orientation calculation), image filtering, and descriptor calculation.

The feature extraction block is exercised by both the left and right images. Features and descriptors of a image pair are used by the data association block to get matched feature pairs. The depth of the landmark associated with the feature is calculated according to the disparity of the matched feature. Depths and associated features are then used to initialize a 3D feature map and poses, which are implemented on the embedded processor. Then, the estimated 3D features and camera poses are transferred to the back-end hardware. 
	
\textbf{Back-end Implementation}.
We profile the LM algorithm on data-sets\cite{Agarwal2010Bundle}. Jacobian updates (JU), Schur elimination (SE), cost function computation (CC) are the top three time-consuming parts, of which computations account for 51\%, 29.8\% and 10.45\% of total time. 

We propose a hardware-software co-design \cite{liu2020pi} in which the time-consuming parts JU, SE and CC are accelerated by the hardware; Cholesky factorization (CFS) and gain ratio evaluation (GRE) are implemented by the software. DMA is used for data exchange between the hardware and software. The hardware optimizes camera poses and feature maps in a sliding window manner. The maximum window size, i.e. the number of camera poses, is 50.

\subsection{Evaluation Results}
We evaluates the front-end and back-end respectively on a Zynq Ultrascale+ ZCU102, which integrates a quad-core ARM Cortex-A53 CPU with an FPGA on the same chip. 

\textbf{Resource consumption.} Table \ref{tab:resource} summarizes the hardware resource consumption of the front-end and back-end. Front-end dominates the resource consumption. The front-end and back-end consume too much BRAM to be fitted together in one Zynq FPGA chip. To solve this problem, we take a holistic approach to optimize the on-chip memory and logic of the SLAM system (will appear in \cite{gan2020hpca}).

\textbf{Performance.} 
Table \ref{tab:perf} shows the performance comparison between front-end and back-end FPGA and Intel CPU. Compared with Intel CPU, front-end and back-end FPGA is 4.8 and 9.8 times faster, and reduce 96\% and 92\% power consumption.

\begin{table}[]
	\begin{threeparttable}[b]
	\caption{Performance comparison between FPGA and Intel CPU.}
	\label{tab:perf}
	\begin{tabular}{ccccc}
	\hline
            & \begin{tabular}[c]{@{}c@{}}Front-end\\ FPGA\end{tabular} & \begin{tabular}[c]{@{}c@{}}Front-end\\ CPU\end{tabular} & \begin{tabular}[c]{@{}c@{}}Back-end\\ FPGA\tnote{a}\end{tabular} & \begin{tabular}[c]{@{}c@{}}Back-end\\ CPU\end{tabular} \\ \hline
Latency(ms) & 19.7                                                     & 95                                                      & 7.6                                                     & 75                                                     \\
Power(W)    & 2.3                                                      & $\sim$65                                                & 5.5                                                     & $\sim$65                                               \\ \hline
	\end{tabular}
	\begin{tablenotes}
		\item[a] Back-end FPGA consists of JU, SE and CC.
	\end{tablenotes}
	\end{threeparttable}
\vspace{-5mm}
\end{table}

\section{Summary and Discussion}
This paper is a concise summary of PerceptIn's efforts on designing the on-vehicle computing system for our commercial autonomous vehicles. By thoroughly characterize the workloads, we adopt a server+FPGA heterogeneous platform, in which localization front-end is offloaded to FPGA. By offloading, the perception latency improves by 1.6$\times$. In retrospective, PerceptIn shipped its products globally, but delayed by six months because PerceptIn took an R\&D detour to explore using existing mobile SoCs for autonomous driving workloads. If PerceptIn had chosen to develop its proprietary computing system initially, PerceptIn would have greatly widened its moat, and enlarging its edge over competitions. The root cause behind this problem was that we didn't have any good tools to help us quickly explore the design space, and to identify the right design decision. In general, we feel that the whole community of autonomous machine computing system development, regardless whether it is academia or industry, are exploring in an enormous design space, and we are limited by the tools available to us.  As a next step of our research, we will focus on developing the necessary scenario simulators, benchmarks, data sets, and SoC simulators needed to accelerate the design flow for autonomous machine computing systems.



\bibliographystyle{ACM-Reference-Format}
\bibliography{ref}

\begin{thebibliography}{10}
\providecommand{\url}[1]{#1}
\csname url@samestyle\endcsname
\providecommand{\newblock}{\relax}
\providecommand{\bibinfo}[2]{#2}
\providecommand{\BIBentrySTDinterwordspacing}{\spaceskip=0pt\relax}
\providecommand{\BIBentryALTinterwordstretchfactor}{4}
\providecommand{\BIBentryALTinterwordspacing}{\spaceskip=\fontdimen2\font plus
\BIBentryALTinterwordstretchfactor\fontdimen3\font minus
  \fontdimen4\font\relax}
\providecommand{\BIBforeignlanguage}[2]{{%
\expandafter\ifx\csname l@#1\endcsname\relax
\typeout{** WARNING: IEEEtran.bst: No hyphenation pattern has been}%
\typeout{** loaded for the language `#1'. Using the pattern for}%
\typeout{** the default language instead.}%
\else
\language=\csname l@#1\endcsname
\fi
#2}}
\providecommand{\BIBdecl}{\relax}
\BIBdecl

\bibitem{liu2020critical}
S.~Liu, ``Critical business decision making for technology startups: A
  perceptin case study,'' \emph{IEEE Engineering Management Review}, 2020.

\bibitem{liu2020autonomous}
S.~Liu and J.-L. Gaudiot, ``Autonomous vehicles lite self-driving technologies
  should start small, go slow,'' \emph{IEEE Spectrum}, vol.~57, no.~3, pp.
  36--49, 2020.

\bibitem{santos2011summary}
A.~Santos, N.~McGuckin, H.~Y. Nakamoto, D.~Gray, S.~Liss \emph{et~al.},
  ``Summary of travel trends: 2009 national household travel survey,'' United
  States. Federal Highway Administration, Tech. Rep., 2011.

\bibitem{liu2020engineering}
S.~Liu, \emph{Engineering Autonomous Vehicles and Robots: The DragonFly
  Modular-based Approach}.\hskip 1em plus 0.5em minus 0.4em\relax John Wiley \&
  Sons, 2020.

\bibitem{liu2017computer}
S.~Liu, J.~Tang, Z.~Zhang, and J.-L. Gaudiot, ``Computer architectures for
  autonomous driving,'' \emph{Computer}, vol.~50, no.~8, pp. 18--25, 2017.

\bibitem{liu2019edge}
S.~Liu, L.~Liu, J.~Tang, B.~Yu, Y.~Wang, and W.~Shi, ``Edge computing for
  autonomous driving: Opportunities and challenges,'' \emph{Proceedings of the
  IEEE}, vol. 107, no.~8, pp. 1697--1716, 2019.

\bibitem{liu2020pirt}
L.~Liu, J.~Tang, S.~Liu, B.~Yu, J.-L. Gaudiot, and Y.~Xie, ``$\pi$-rt: A
  runtime framework to enable energy-efficient real-time robotic vision
  applications on heterogeneous architectures,'' \emph{Computer}, vol.~54,
  2021.

\bibitem{fang2018dragonfly+}
W.~Fang, Y.~Zhang, B.~Yu, and S.~Liu, ``Dragonfly+: Fpga-based quad-camera
  visual slam system for autonomous vehicles,'' \emph{Proc. IEEE HotChips},
  p.~1, 2018.

\bibitem{fang2017fpga}
------, ``Fpga-based orb feature extraction for real-time visual slam,'' in
  \emph{2017 International Conference on Field Programmable Technology
  (ICFPT)}.\hskip 1em plus 0.5em minus 0.4em\relax IEEE, 2017, pp. 275--278.

\bibitem{liu2020pi}
Q.~Liu, S.~Qin, B.~Yu, J.~Tang, and S.~Liu, ``$\pi$-ba: Bundle adjustment
  hardware accelerator based on distribution of 3d-point observations,''
  \emph{IEEE Transactions on Computers}, 2020.

\bibitem{tang2018pi}
J.~Tang, B.~Yu, S.~Liu, Z.~Zhang, W.~Fang, and Y.~Zhang, ``$\pi$-soc:
  Heterogeneous soc architecture for visual inertial slam applications,'' in
  \emph{2018 IEEE/RSJ International Conference on Intelligent Robots and
  Systems (IROS)}.\hskip 1em plus 0.5em minus 0.4em\relax IEEE, 2018, pp.
  8302--8307.

\bibitem{yu2020building}
B.~Yu, W.~Hu, L.~Xu, J.~Tang, S.~Liu, and Y.~Zhu, ``Building the computing
  system for autonomous micromobility vehicles: Design constraints and
  architectural optimizations,'' in \emph{2020 53rd Annual IEEE/ACM
  International Symposium on Microarchitecture (MICRO), IEEE}, 2020.

\bibitem{qin2018vins}
T.~Qin, P.~Li, and S.~Shen, ``Vins-mono: A robust and versatile monocular
  visual-inertial state estimator,'' \emph{IEEE Transactions on Robotics},
  vol.~34, no.~4, pp. 1004--1020, 2018.

\bibitem{sun2018msckf}
K.~{Sun}, K.~{Mohta}, B.~{Pfrommer}, M.~{Watterson}, S.~{Liu}, Y.~{Mulgaonkar},
  C.~J. {Taylor}, and V.~{Kumar}, ``Robust stereo visual inertial odometry for
  fast autonomous flight,'' \emph{IEEE Robotics and Automation Letters},
  vol.~3, no.~2, pp. 965--972, 2018.

\bibitem{szeliski2010computer}
\BIBentryALTinterwordspacing
R.~Szeliski, \emph{Computer Vision: Algorithms and Applications}, ser. Texts in
  Computer Science.\hskip 1em plus 0.5em minus 0.4em\relax Springer London,
  2010. [Online]. Available:
  \url{https://books.google.com/books?id=bXzAlkODwa8C}
\BIBentrySTDinterwordspacing

\bibitem{Elas2010Geiger}
A.~Geiger, M.~Roser, and R.~Urtasun, ``Efficient large-scale stereo matching,''
  in \emph{Proceedings of the 10th Asian Conference on Computer Vision}, 2010.

\bibitem{feng2019stereo}
Y.~Feng, P.~Whatmough, and Y.~Zhu, ``Asv: Accelerated stereo vision system,''
  in \emph{Proceedings of the 52nd Annual IEEE/ACM International Symposium on
  Microarchitecture}, ser. MICRO '52, 2019, p. 643–656.

\bibitem{henriques2014high}
J.~F. Henriques, R.~Caseiro, P.~Martins, and J.~Batista, ``High-speed tracking
  with kernelized correlation filters,'' \emph{IEEE transactions on pattern
  analysis and machine intelligence}, vol.~37, no.~3, pp. 583--596, 2014.

\bibitem{kelly2013mobile}
A.~Kelly, \emph{Mobile robotics: mathematics, models, and methods}.\hskip 1em
  plus 0.5em minus 0.4em\relax Cambridge University Press, 2013.

\bibitem{redmon2018yolov3}
J.~Redmon and A.~Farhadi, ``Yolov3: An incremental improvement,'' 2018.

\bibitem{mur2017orb}
R.~Mur-Artal and J.~D. Tard{\'o}s, ``Orb-slam2: An open-source slam system for
  monocular, stereo, and rgb-d cameras,'' \emph{IEEE Transactions on Robotics},
  vol.~33, no.~5, pp. 1255--1262, 2017.

\bibitem{kalibr2013}
P.~{Furgale}, J.~{Rehder}, and R.~{Siegwart}, ``Unified temporal and spatial
  calibration for multi-sensor systems,'' in \emph{2013 IEEE/RSJ International
  Conference on Intelligent Robots and Systems}, 2013, pp. 1280--1286.

\bibitem{Triggs1999Bundle}
B.~Triggs, P.~F. McLauchlan, R.~I. Hartley, and A.~W. Fitzgibbon, \emph{Bundle
  adjustment—a modern synthesis}.\hskip 1em plus 0.5em minus 0.4em\relax
  Springer Berlin Heidelberg, 1999.

\bibitem{Lourakis2005Is}
M.~Lourakis and A.~A. Argyros, ``Is {L}evenberg-{M}arquardt the most efficient
  optimization algorithm for implementing bundle adjustment?'' in
  \emph{Computer Vision, 2005. ICCV 2005. Tenth IEEE International Conference
  on}, vol.~2.\hskip 1em plus 0.5em minus 0.4em\relax IEEE, 2005, pp.
  1526--1531.

\bibitem{Agarwal2010Bundle}
S.~Agarwal, N.~Snavely, S.~M. Seitz, and R.~Szeliski, ``Bundle adjustment in
  the large,'' in \emph{European Conference on Computer Vision}, 2010, pp.
  29--42.

\bibitem{gan2020hpca}
Y.~Gan, B.~Yu, B.~Tian, L.~Xu, W.~Hu, J.~Tang, S.~Liu, and Y.~Zhu, ``Eudoxus:
  Characterizing and accelerating localization in autonomous machines,'' (to
  appear HPCA2021).

\end{thebibliography}


\end{document}